\documentclass[letterpaper, 10 pt, conference]{ieeeconf} 
\IEEEoverridecommandlockouts                        
\usepackage{graphics} 
\usepackage{epsfig} 
\usepackage{mathptmx} 
\usepackage{times} 
\usepackage{amsmath} 
\usepackage{amssymb}  
\usepackage{multirow}
\usepackage{graphicx}
\usepackage{amsmath}
\usepackage{amssymb}
\usepackage{booktabs}
\usepackage{multirow}
\usepackage{color}
\usepackage[dvipsnames]{xcolor}
\usepackage{colortbl}
\usepackage{amsfonts}
\usepackage{accents}
\usepackage{mathtools}
\usepackage{adjustbox}
\usepackage{bbding}
\usepackage{wrapfig}
\usepackage[colorlinks,linkcolor=red,anchorcolor=blue,citecolor=green]{hyperref}
\usepackage{cite}

\newcommand{\orcid}[1]{\href{https://orcid.org/#1}{\includegraphics[width=10pt]{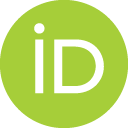}}}

\title{\LARGE \bf
Motion-to-Matching: A Mixed Paradigm for 3D Single Object Tracking
}

\author{Zhiheng Li\orcid{0000-0002-1477-2066}, Yu Lin\orcid{0009-0008-7820-4835}, Yubo Cui\orcid{0000-0001-5302-0484}, Shuo Li\orcid{0009-0005-9318-0219}, and Zheng Fang*\orcid{0000-0003-3887-3141}
\thanks{This work was supported in part by National Natural Science Foundation of China under Grants 62073066 and U20A20197, in part by Fundamental Research Funds for Central Universities under Grant N2226001, and in part by 111 Project under Grant B16009. (\textit{Corresponding author: Zheng Fang})}
\thanks{The authors are all with the Faculty of Robot Science and Engineering, Northeastern University, Shenyang 110819, China. Zhiheng Li and Zheng Fang are also with the National Frontiers Science Center for Industrial Intelligence and Systems Optimization, Northeastern University, Shenyang 110819, China and also with Key Laboratory of Data Analytics and Optimization for Smart Industry, Ministry of Education, Northeastern University, Shenyang 110819, China. (e-mail: fangzheng@mail.neu.edu.cn)}
}

\begin{document}

\maketitle
\thispagestyle{empty}
\pagestyle{empty}

\begin{abstract}
3D single object tracking with LiDAR points is an important task in the computer vision field. 
Previous methods usually adopt the matching-based or motion-centric paradigms to estimate the current target status. However, the former is sensitive to the similar distractors and the sparseness of point clouds due to relying on appearance matching, while the latter usually focuses on short-term motion clues (eg. two frames) and ignores the long-term motion pattern of target. To address these issues, we propose a mixed paradigm with two stages, named MTM-Tracker, which combines motion modeling with feature matching into a single network. Specifically, in the first stage, we exploit the continuous historical boxes as motion prior and propose an encoder-decoder structure to locate target coarsely. Then, in the second stage, we introduce a feature interaction module to extract motion-aware features from consecutive point clouds and match them to refine target movement as well as regress other target states. Extensive experiments validate that our paradigm achieves competitive performance on large-scale datasets (70.9\% in KITTI and 51.70\% in NuScenes). The code will be open soon at \url{https://github.com/LeoZhiheng/MTM-Tracker.git}.
\end{abstract}

\section{INTRODUCTION}
\label{sec:introduction}
\begin{figure}[t]
\centering
\includegraphics[width=7.6cm, height=7.4cm]{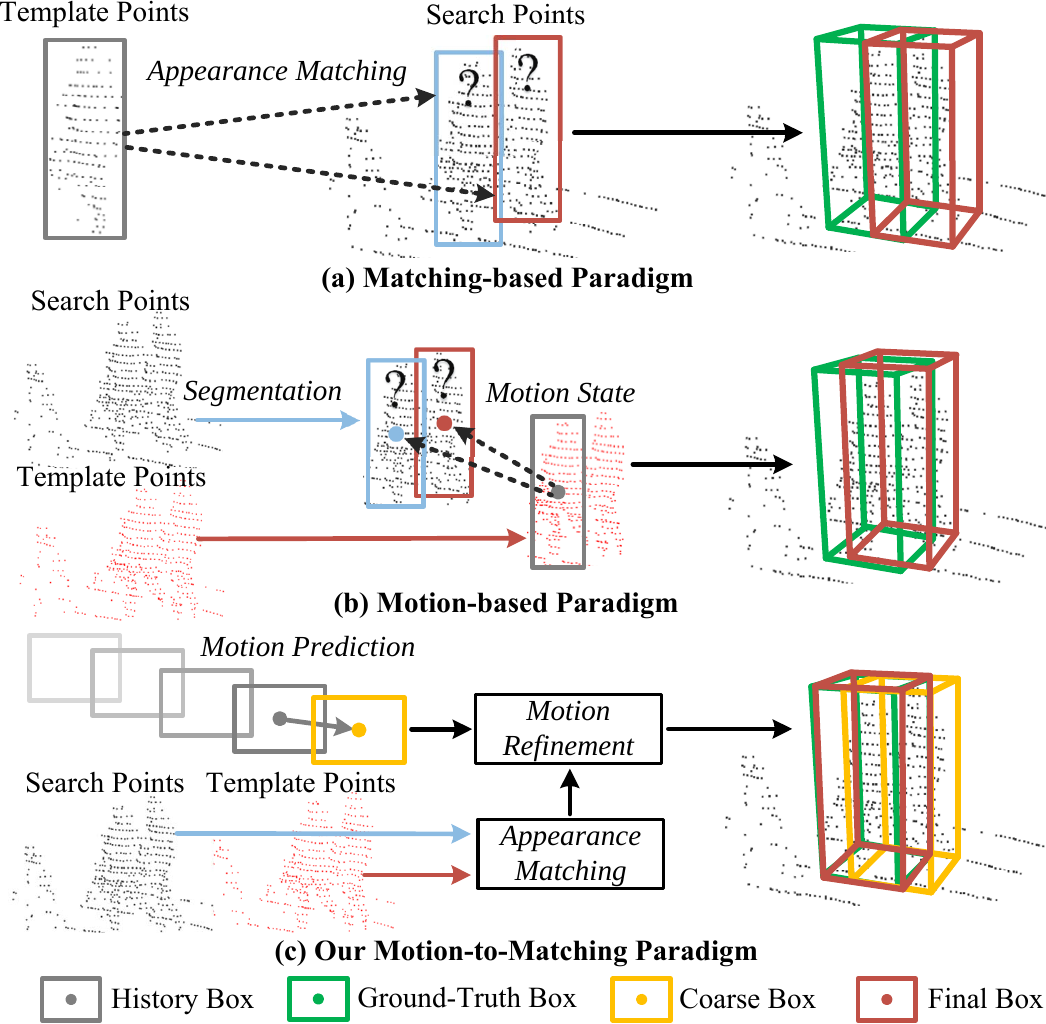}
\vspace{-0.05in}
\caption{\textbf{Comparison of 3D single object tracker paradigm.} (a) Matching-based paradigm exploits appearance matching to locate target. (b) Motion-centric paradigm utilizes segmented points to predict the relative motion. (c) Our Motion-to-Matching paradigm adopts historical boxes to locate target roughly and uses matching feature to refine it as well as regress target box.}
\label{fig:Compare Framework}
\vspace{-0.2in}
\end{figure}

Recently, 3D Single Object Tracking (SOT) has received widespread attention in 3D computer vision and has broad application prospect in autonomous driving and mobile robot.
Although existing 3D SOT approaches~\cite{P2B,BAT,PTT,lttr,v2b,C2FT,MLSET} have achieved promising results, it is noteworthy that most of them inherit the Siamese structure from 2D SOT~\cite{SiamFC,SiamRPN,SiamMask} and regard 3D tracking as a matching problem (in Fig.~\ref{fig:Compare Framework}(a)), which ignores the difference between image and point clouds. Specifically, for 2D SOT, the dense and rich image textures make target matching in consecutive frames relatively simple. However, limited by sparse and textureless nature of point clouds, the 3D matching-based trackers are sensitive to similar objects and prone to fail when tracking occluded target.

To break through the limitations of the matching paradigm, some works try to exploit motion or temporal information. Specially, as shown in Fig.~\ref{fig:Compare Framework}(b), M$^2$-Track~\cite{beyond} proposes a motion-centric paradigm which first segments two consecutive point clouds and then estimates the relative displacement of target based on foreground points. 
Besides, to use multi-view target appearance, TAT~\cite{TAT} selects some high-quality templates from the previous frames and propagates temporal features to the current frame. 
Nevertheless, both of them have some shortcomings: 1) M$^2$-Track~\cite{beyond} utilizes the short-term motion clues between two frames but ignores the long-term motion pattern presented in historical frames. Meanwhile, erroneous segmentation makes it hard to identify target from distractors. 
2) In the case of using TAT~\cite{TAT} to track distant target, the point clouds are usually so sparse that it is difficult to match target accurately even integrating incomplete points from previous frames.

Different from aggregating multi-template features or estimating target motion via two adjacent point cloud frames, we observe that the center and corners of historical boxes could also describe the target's key states (eg. consecutive location and orientation), which are usually ignored before. Therefore, these historical boxes could be utilized as a strong prior to model the motion pattern of target and realize rough localization. Meanwhile, owing to only processing a series of boxes, we could avoid extensive computational consumption, compared to dealing with numerous points. However, despite being simple and efficient, it inevitably accumulates location errors due to some low-quality predicted boxes in long-term tracking. Fortunately, benefiting from the accurate geometry information provided by the point clouds, it is easy to further eliminate errors via feature matching.

Based on the above viewpoints, we propose a mixed tracking paradigm (in Fig.~\ref{fig:Compare Framework}(c)), called MTM-Tracker (\textbf{M}otion-\textbf{T}o-\textbf{M}atching Tracker), which integrates the motion modeling and feature matching in the two-stage framework. 
\textbf{In the first stage}, the core problem is to estimate target motion via a set of historical boxes. A naive approach is to adopt LSTM~\cite{lstm} or GRU~\cite{gru} to aggregate previous motion clues in the hidden features. However, if any previous box contains significant noise, the predicted target motion would be largely disrupted. Therefore, we present a \textit{Box Motion Predictor} (BMP) based on Transformer encoder-decoder module to adaptively learn spatio-temporal transformation of target and suppress noise by attention mechanism. 
After that, \textbf{in the second stage}, we aim to refine target location through appearance matching of point clouds. We first extract geometry features from sparse points and then introduce a \textit{Reciprocating Interaction Module} (RIM) which not only propagates target-specific features in up-down and bottom-top streams but also constructs location associations among the search and template regions. In this way, our tracker could perceive the target displacement in feature patches.
Moreover, we introduce an \textit{Iterative Refinement Module} (IRM) to match motion-aware features between two frames and obtain a correlation volume, which is utilized to refine target motion layer by layer. Finally, by gathering the refined motion and regressing other attributes, we could realize robust tracking for a special object.

Overall, our contributions are as follows: \textbf{1)} A novel mixed paradigm for 3D SOT, which combines the long-term motion modeling with appearance matching; \textbf{2)} We propose a two-stage pipeline that contains three proposed modules to realize motion prediction, feature interaction and motion refinement; \textbf{3)} Experiments on the KITTI and NuScenes datasets show that our method achieves competitive performance, and ablation studies verify the effectiveness of the proposed modules.

\section{RELATED WORK}
\subsection{2D Single Object Tracking} 
For decades, 2D single object tracking (SOT) has become a fundamental task in image perception and aims to estimate the state of an arbitrary target in the video sequences based on initial status. With the rapid development of deep learning, most 2D SOT trackers~\cite{SiamFC,SiamRPN,siamcar,siameseban} employ the Siamese network for feature matching and obtain a similarity map to locate target. For example, pioneering approach SiamFC~\cite{SiamFC} used weight-shared network to extract features and adopted cross-correlation layer to highlight target on response map. However, due to missing bounding box regression, SiamFC cannot achieve satisfactory results. Thus, SiamRPN~\cite{SiamRPN} introduced classification and proposal regression branches to improve the target localization. Additionally, SiamMask~\cite{SiamMask} performed both target tracking and segmentation through a unified network. SiamCAR~\cite{siamcar} and SiamBAN~\cite{siameseban} adopted pixel-wise regression manner to avoid setting the numerous anchors in advance. In recent year, encouraged by impressive performance of Transformer~\cite{Transformer}, numerous works~\cite{transt,transformermeetstracker} attempt to construct semantic relation of target between two frames through attention mechanism. Specially, TransT~\cite{transt} replaced similarity calculation with Transformer to propagate target feature to the search region. TrDiMP~\cite{transformermeetstracker} introduced Transformer structure to convey rich temporal clues across frames. However, due to lacking depth information in RGB images, it is non-trivial to use 2D methods to track target in 3D space. 

\subsection{3D Single Object Tracking}
Due to the emergence of LiDAR sensor, many works~\cite{SC3D,P2B,BAT,PTT,lttr,pttr,v2b,C2FT,MLSET} refer to Siamese structures and exploit point clouds to realize 3D tracking. Initially, SC3D~\cite{SC3D} executed Kalman Filter to generate bunches of candidate shapes and chose the best candidate by the Siamese network. However, SC3D cannot be end-to-end trained and consumes much time due to hand-crafted filter. To solve the above issues, P2B~\cite{P2B} generated target-specific features to predict potential centers and utilized VoteNet~\cite{votenet} to predict accurate target proposals in an end-to-end manner. After that, some works like~\cite{BAT},~\cite{v2b} follow the P2B pipeline, but the difference is that BAT~\cite{BAT} exploited box-aware features to strengthen similarity learning. And the V2B~\cite{v2b} regressed target center on the dense bird’s eye view (BEV) feature through Voxel-to-BEV network. At present, more methods~\cite{PTT,pttr,SMAT,stnet,lttr} show excellent performance by using attention mechanism. Specifically, PTT~\cite{PTT} embeded Transformer blocks into center voting and proposal generation stages. LTTR~\cite{lttr} proposed Transformer encoder-decoder structure to learn semantic relations among different regions. PTTR~\cite{pttr} adopted attention-based feature matching to enhance point feature and integrated target clues.

Although matching-based methods have achieved success in certain, they ignore that tracking is a continuous process in space-time, and break motion association among historical frames. Thus, these methods are sensitive to distractors in the surroundings. Focusing on this problem, M$^2$-Track~\cite{beyond} proposed a motion-based method that modelled relative motion of target instead of feature matching. However, due to only considering motion between two frames, M$^2$-Track still does not fully exploit the temporal information. On the contrary, although recent TAT~\cite{TAT} exploited multi-template features in historical frames, TAT did not consider their motion relation. Therefore, in this paper, we propose a new paradigm which exploits historical boxes as strong prior to model continuous target motion and further refine target location by geometry information from point clouds.

\begin{figure*}[ht]
\centering
\vspace{0.1in}
\includegraphics[width=16.5cm,height=5.2cm]{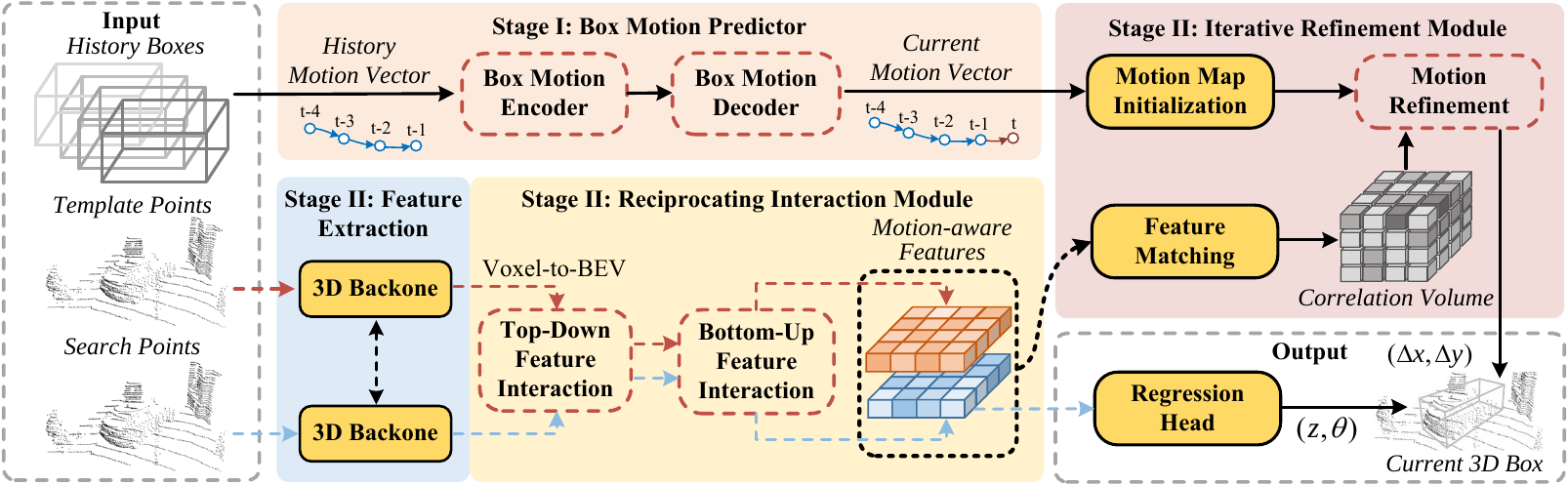}
 \vspace{-0.05in}
\caption{\textbf{The overall architecture of MTM-Tracker.} At the 1$^{st}$ stage, for a set of historical boxes, MTM-Tracker uses them to predict a coarse box motion via \textit{Box Motion Predictor}. At the 2$^{nd}$ stage, we extract geometry features from consecutive point clouds and perform feature interaction by \textit{Reciprocating Interaction Module}. Finally, we exploit motion-aware features to refine target motion in \textit{Iterative Refinement Module} and regress other target attributes.}
\label{pic: network overview}
 \vspace{-0.15in}
\end{figure*}

\begin{figure}[t]
\centering
\includegraphics[width=\linewidth]{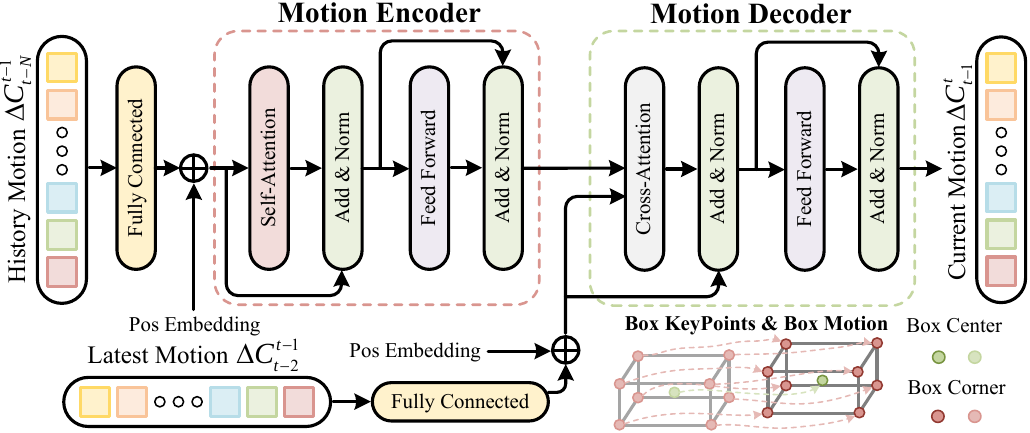}
\caption{\textbf{The structure of Box Motion Predictor (BMP).} BMP utilizes historical box keypoints to predict target motion via encoder-decoder stream.}
\vspace{-0.2in}
\label{fig:Data-Driven Filter}
\end{figure}

\section{METHODOLOGY}
\subsection{Problem Statement}
Unlike multi-object tracking, SOT only involves one target throughout tracking process and does not rely on detection methods to discover new targets. Thus, given a 3D bounding box $B_{0}$ of target in the first frame, most 3D SOT trackers locate target in subsequent frames according to the last predicted box $B_{t-1}$ and two continuous point clouds $(P_{t}, P_{t-1})$. Additionally, the size of target is usually assumed to remain unchanged~\cite{P2B,BAT}, thus we only require to predict the center $(x, y, z)$ and orientation $\theta$ of target.

\subsection{Overall Architecture}
Different from previous work only using $B_{t-1}$, we consider historical boxes $B_{t-N}^{t-1}=[B_{t-N},...,B_{t-1}]$ in the $N$ previous frames as prior motion information, which can guide tracker to locate target from sparse LiDAR data. As shown in Fig.~\ref{pic: network overview}, we propose two-stage network MTM-Tracker, which exploits historical boxes to estimate coarse target motion and utilizes geometry information from point cloud to estimate the target box $B_{t}$. 
Specifically, in the first stage, given a historical box sequence, we introduce a Box Motion Predictor (BMP) which leverages the box keypoints to estimate target motion roughly. 
Then, in the second stage, we present a Reciprocating Interaction Module (RIM) to learn spatial relations among search and template regions through top-down and bottom-up feature interaction. 
Moreover, to refine the coarse motion in BMP, we propose an Iterative Refinement Module (IRM) to initialize a motion map and incrementally refine it by geometry correlation. We will detail each module in the following subsections.

\subsection{Stage I: Box Motion Prediction} 
\label{sec:Data-Driven Filter}
\noindent \textbf{Box Keypoints Representation.}
We observe that the spatial status of any previous box can be efficiently determined by its eight corners and one center point. Thus, we arrange them in a predefined order to form a keypoint vector $C\in \mathbb{R}^{9K}$, where $K$ represents the coordinate of keypoints. And a group of historical boxes can be formulated as $C_{t-N}^{t-1}=[C_{t-N},...,C_{t-1}]$. 
After that, we calculate the keypoint offsets for every adjacent box, so that the relative motion in each time step can be denoted as $\Delta C_{t-N}^{t-1} = [C_{t-N} - C_{t-N+1},..., C_{t-2} - C_{t-1}]$.

\noindent \textbf{Box Motion Predictor.}
After obtaining historical movement corresponding to each keypoint, our goal is to predict current motion through these spatial-temporal information. However, due to some low-quality predicted boxes, historical keypoints may deviate from the real position and destroy overall motion prediction. 
Inspired by Transformer~\cite{Transformer}, we propose a Box Motion Predictor based on encoder-decoder structure, aiming to construct global association among keypoints and suppress noise by learnable weights. The detail of BMP is illustrated in Fig.~\ref{fig:Data-Driven Filter}. Specifically, we first apply fully connected layer to the historical motion $\Delta C_{t-N}^{t-1} \in \mathbb{R}^{(N-1) \times 9K}$ and add sinusoidal positional embedding~\cite{Transformer} to generate a motion token $E^{m}\in \mathbb{R}^{(N-1) \times D}$ for $N$ time steps, where $D$ is channel dimension. Then, we feed $E^{m}$ to Transformer encoder to learn global-range motion feature $\tilde E^{m}$. The process can be formulated as:
\begin{equation}
Q^{m} = E^{m}W^{q}, K^{m} = E^{m}W^{k}, V^{m} = E^{m}W^{v}
\end{equation}
\begin{equation}
\hat E^{m}=\text{LN}\big(\text{MHA}(Q^{m},K^{m},V^{m})+Q^{m}\big)
\end{equation}
\begin{equation}
\tilde E^{m}=\text{LN}\big(\text{FFN}(\hat E^{m})+\hat E^{m}\big)
\end{equation}
where $Q,K,V$ are query, key and value embedding obtained by projection matrices $W^{q}, W^{k}, W^{v}$. And the MHA, LN, FFN represent multi-head attention, layer normalization and feed-forward network respectively. 
Furthermore, for Transformer decoder, we convert the latest motion vector $\Delta C_{t-2}^{t-1}$ to motion token $E^{n}$ and combine it with $\tilde E^{m}$ to decode current motion $\Delta C_{t-1}^{t}$. Note that the decoder is similar to the encoder except that we generate the query embedding through $E^{n}$ and project $\tilde E^{m}$ to key and value embedding respectively. Finally, through averaging the coordinate of keypoint offsets $\Delta C_{t-1}^{t}$, we obtain coarse motion vector $V_{t-1}^{t}=\{\Delta x_{t-1}^{t}, \Delta y_{t-1}^{t}, \Delta z_{t-1}^{t}\}$ for target center and can locate target in the current frame.

\begin{figure*}[t!]
	\centering
 \vspace{0.1in}	\includegraphics[width=16.5cm,height=8.3cm]{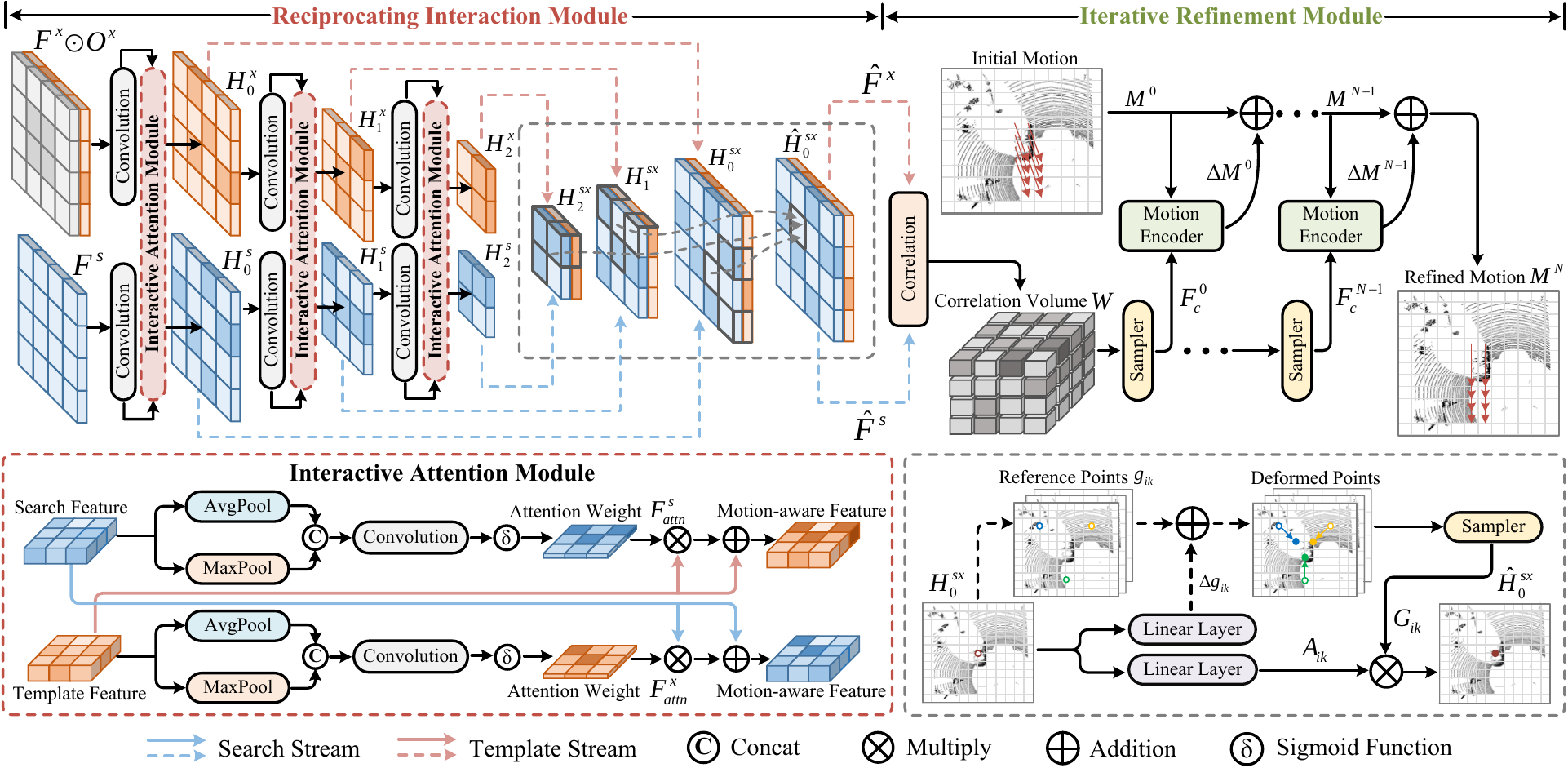}
 \vspace{-0.1in}
	\caption{\textbf{Illustration of the proposed Reciprocating Interaction Module (RIM) and Iterative Refinement Module (IRM).} The RIM consists of two parts, top-down and bottom-up streams. The former expands the receptive field while performing feature interaction, then the latter aggregates and reweights multi-scale features via deformable mechanism. The IRM exploits motion-aware features to generate correlation volume and use it to refine target motion.}
	\label{fig:RIM and IRM}
 \vspace{-0.2in}
\end{figure*}

\subsection{Stage II: Feature Interaction and Motion Refinement} 
\label{sec:Stage II} 
Although stage I can obtain a rough estimation, it is still difficult to reflect the current state of target due to completely depending on historical boxes information. To have a more accurate prediction, we further exploit geometry information from point cloud at time $t$ and $t-1$ to refine target location.
\subsubsection{Point Feature Extraction} 
Firstly, we crop out template points with the same size as search branch to fully use contextual information surrounding target. Then, following~\cite{lttr,SMAT}, we divide point clouds into uniformly sized voxels and apply a weight-shared 3D sparse convolution backbone~\cite{second} to aggregate local geometry features. To get a dense representation, we merge the z-axis and feature channel for voxel feature and get 2D BEV feature $F \in \mathbb{R}^{H\times L\times D}$ to feed feature interaction module in the following network, where $H, L, D$ are the height, width and channel dimensions respectively.

\subsubsection{Reciprocating Interaction Module} 
\label{sec:Reciprocating Feature Interaction}
In order to perceive target motion, we propose Reciprocating Interaction Module (RIM) to construct motion relation for search and template regions through top-down and bottom-up feature interaction.

\noindent\textbf{Top-Down Interaction.}
To exploit prior information in the template frame, we first generate mask $O^{x}\in \mathbb{R}^{H\times L\times 1}$ which sets 0 or 1 to denote feature grid whether occupied by the target. Then, we concatenate template feature $F^{x}$ with mask $O^{x}$ along feature channel and perform multi-layer interaction with search stream. For each layer, we adopt a convolution block to downsample features with downsampling rates of $2^{i}$, where $i\in \{0, 1, 2\}$ in the corresponding layer. Then, we design an interactive attention module (IAM) to learn target-specific associations. As displayed in Fig.~\ref{fig:RIM and IRM}, for the template and search features, we utilize pooling operation to get global features and generate attention weight by following equation:
\begin{equation}
F_{attn} = \sigma \Big(\text{Conv2D}\big(\text{AvgPool}(F)\odot \text{MaxPool}(F)\big)\Big)
\end{equation}
where $\odot$ means concatenation operation along channel dimension. The AvgPool and MaxPool denote average pooling and max pooling respectively, and $\sigma$ represents the sigmoid function. Then, $F^{x}_{attn}$ and $F^{s}_{attn}$ are utilized to enhance target-specific feature in $F^{x}$ and $F^{s}$ by cross manner:
\begin{equation}
H^{s} = (F^{x}_{attn} \otimes F^{s}) + F^{s}, \ 
H^{x} = (F^{s}_{attn} \otimes F^{x}) + F^{x}
\end{equation}
where $\otimes$ denotes the multiply operation. In this way, we can guide the network to learn spatio-temporal relation of target in consecutive frames. Meanwhile, by repeating the above steps, we can obtain multi-scale feature pairs $\{H_{i}^{s},H_{i}^{x}\}$ with specific receptive fields. Thus, these features can capture the movements for targets with different speeds and sizes.

\noindent \textbf{Bottom-Up Interaction.}
Inspired by attention mechanism, we apply the deformable attention~\cite{DeformableDETR} to align multi-scale features and aggregate target information. Specifically, we concatenate each feature pair $\{H_{i}^{s},H_{i}^{x}\}$ and generate temporal features $H^{sx}_{i} \in \mathbb{R}^{H\times L\times 2D}$. After that, we set the $K$ reference points $g_{ik}$ to each scale of $H^{sx}_{i}$ and use two linear layers to generate $K$ sampling offsets $\Delta g_{ik}$ and attention weights $A_{ik}$. Then, sampling offsets $\Delta g_{ik}$ are added to the coordinates $g_{ik}$ of reference points to sample reference features $G_{ik}$ from $H^{sx}_{i}$ by bilinear interpolation. Finally, the original feature $H^{sx}_{i}$ can be aggregated by reference features $G_{ik}$ with corresponding weights $A_{ik}$ and obtain refined feature $\hat H^{sx}_{i}$. The process could be formulated as follows:
\begin{equation}
A_{ik} = \text{Linear}(H^{sx}_{i}), \ \Delta g_{ik} = \text{Linear}(H^{sx}_{i})
\end{equation}
\begin{equation}
G_{ik} = S(H^{sx}_{i}, g_{ik} + \Delta g_{ik})
\end{equation}
\begin{equation}
\hat H_{i}^{sx} = \sum_{l=1}^{L} W_l (\sum_{i=1}^{I} \sum_{k=1}^{K} A_{lik} \cdot G_{lik})     
\end{equation}
where $S(·\cdot·)$ and $W_l$ mean bilinear sampling and learnable weights. $L,I,K$ are the number of attention heads, scale layers, and reference points respectively. For output features $\hat H_{i}^{sx}$, we pick out the high resolution $\hat H_{0}^{sx}$ and further split it into $\hat F^{x} \in \mathbb{R}^{H\times L\times D} $ and $\hat F^{s} \in \mathbb{R}^{H\times L\times D}$ based on concatenation order for search and template features.
\subsubsection{Iterative Motion Refinement} 
\label{sec:Iterative Motion Refinement}
Inspired by optical flow estimation~\cite{raft}, we propose an Iterative Refinement Module (IRM) to refine coarse motion $V_{t-1}^{t}$ through dense geometry features $\hat F^{x}$ and $\hat F^{s}$, which could become a bridge to connect motion and matching paradigms. The details are as follows:

\begin{table}[t!]
\renewcommand\tabcolsep{4.5pt}
\begin{center}
\vspace{0.1in}
\caption{Performance comparison on the KITTI dataset. Mean performance is weighted by the number of frames.}\label{tab:kitti result}
		\begin{tabular}{c|c|ccccc}
			\toprule[.05cm]
			& Category & Car   & Pedestrian & Van & Cyclist &Mean\\
			& Frame &\textit{6,424} &\textit{6,088} &\textit{1,248} &\textit{308} &\textit{14,068} \\
			\hline
			\hline
                \rule{0pt}{8pt}
			\multirow{15}*{\rotatebox{90}{\textit{Success}}}
                & SC3D~\cite{SC3D} &41.3 & 18.2 &40.4 &41.5 &31.2 \\
			& P2B~\cite{P2B} &56.2 &28.7 &40.8 &32.1 &42.4 \\
			& PTT~\cite{PTT} &67.8 &44.9 &43.6 &37.2 &55.1 \\
			& BAT~\cite{BAT} &60.5 &42.1 &52.4 &33.7 &51.2 \\
			& LTTR~\cite{lttr} &65.0 &33.2 &35.8 &66.2 &48.7 \\
			& V2B~\cite{v2b} &70.5 &48.3 &50.1 &40.8 &58.4 \\
                & C2FT~\cite{C2FT} & 67.0 & 48.6 & 53.4 & 38.0 & 57.2\\
                & MLSET~\cite{MLSET} &69.7 &50.7 &55.2 &41.0 &59.6\\
			& PTTR~\cite{pttr} &65.2 &50.9 &52.5 &65.1 &57.9 \\
                & SMAT~\cite{SMAT} &71.9 &52.1 &41.4 &61.2 &60.4\\
                & STNet~\cite{stnet} &72.1 &49.9 &58.0 &73.5 &61.3 \\
                & TAT~\cite{TAT} &\underline{72.2} & 57.4 &58.9 &\underline{74.2} &64.7 \\
                & M$^2$-Track~\cite{beyond} &65.5 & 61.5 &53.8 &73.2 &62.9 \\
                & CXTrack~\cite{CXTrack} &69.1 & \underline{67.0} &\underline{60.0} &\underline{74.2} &\underline{67.5} \\
			& \textbf{Ours} &\textbf{73.1} &\textbf{70.4} &\textbf{60.8} &\textbf{76.7} &\textbf{70.9} \\
			\hline
			\hline
                \rule{0pt}{8pt}
			\multirow{15}*{\rotatebox{90}{\textit{Precision}}}
                & SC3D~\cite{SC3D} &57.9 & 37.8 &47.0 &70.4 &48.5 \\
			& P2B~\cite{P2B} &72.8 &49.6 &48.4 &44.7 &60.0 \\
			& PTT~\cite{PTT} &81.8 &72.0 &52.5 &47.3 &74.2 \\
			& BAT~\cite{BAT} &77.7 &70.1 &67.0 &45.4 &72.8 \\
			& LTTR~\cite{lttr} &77.1 &56.8 &45.6 &89.9 &65.8 \\
			& V2B~\cite{v2b} &81.3 &73.5 &58.0 &49.7 &75.2 \\
                & C2FT~\cite{C2FT} & 80.4 & 75.6 & 66.1 & 48.7 & 76.4\\
                & MLSET~\cite{MLSET} & 81.0 & 80.0 & 64.8 & 49.7 & 78.4\\
			& PTTR~\cite{pttr} &77.4 &81.6 &61.8 &90.5 &78.1 \\
                & SMAT~\cite{SMAT} &82.4 &81.5 &53.2 &87.3 &79.5 \\
                & STNet~\cite{stnet} &\underline{84.0} &77.2 &70.6 &93.7 &80.1 \\
                & TAT~\cite{TAT} &83.3 & 84.4 &69.2 &93.9 &82.8 \\
                & M$^2$-Track~\cite{beyond} &80.8 &88.2 &70.7 &93.5 &83.4 \\
                & CXTrack~\cite{CXTrack} &81.6 & \underline{91.5} &\underline{71.8} &\underline{94.3} &\underline{85.3} \\
			& \textbf{Ours} &\textbf{84.5} &\textbf{95.1} &\textbf{74.2} &\textbf{94.6} &\textbf{88.4} \\
			\toprule[.05cm]
		\end{tabular}
	\end{center}
 \vspace{-0.3in}
\end{table}

\noindent \textbf{Motion Initialization.} 
We initialize the motion map $M^{0}=(f^{1}, f^{2}) \in \mathbb{R}^{H\times L\times 2}$ to 0 everywhere, where $f^1$ and $f^2$ means motion components along the X and Y axes. Then, we utilize $\Delta x_{t-1}^{t}$ and $\Delta y_{t-1}^{t}$ in coarse motion vector $V_{t-1}^{t}$ from BMP to fill $f^{1}$ and $f^{2}$ respectively.

\noindent \textbf{Correlation Volume Generation.}
Given dense features $\hat F^{x}$ and $\hat F^{s}$, we employ dot product to them and form a 4D correlation volume $W \in  \mathbb{R}^{H \times L \times H \times L}$, which represents similarity degree for each pixel pair between $\hat F^{x}(u, v)$ and $\hat F^{s}(u, v)$.

\noindent \textbf{Iterative Refinement.}
In this part, our goal is to iteratively update $M^{0}$ in $N$ times and get a refined $M^{N}$. As displayed in Fig.~\ref{fig:RIM and IRM}, for each iteration, we estimate an increment $\Delta M$ and add it to current motion: $M^{n+1} = M^{n} + \Delta M^{n}$. Specifically, based on $M^{n}$, we project each pixel $p = (u, v)$ in $\hat F^{x}$ to its estimated correspondence in $\hat F^{s}: p' = (u + f^{1}(u), v + f^{2}(v))$. After that, we generate coordinate indices within radius r of each pixel $p'$ to sample correlation feature $F_c^{n} \in \mathbb{R}^{W \times L \times r^{2}}$ from volume $W$. Besides, we employ several convolutional blocks to process the correlation feature and combine it with $M^{n}$ to generate $\Delta M^{n}$. The process could be defined as:
\begin{equation}
R = \text{ConvB}\big(\text{ConvB}(F_c^{n}) \odot \text{ConvB}(M^{n})\big) \odot M^{n}
\end{equation}
\begin{equation}
\Delta M^{n} = \text{Conv2D}\big(\text{ConvB}(R)\big)
\end{equation}
where ConvB is the block consisting of 2D convolution and ReLU layer. As a result, we could exploit geometry features to eliminate the estimation bias caused by the first stage as much as possible and get a more accurate motion map $M^{N}$.

\begin{figure*}[t!]
	\centering
 \vspace{0.1in}
\includegraphics[width=17cm,height=5.4cm]{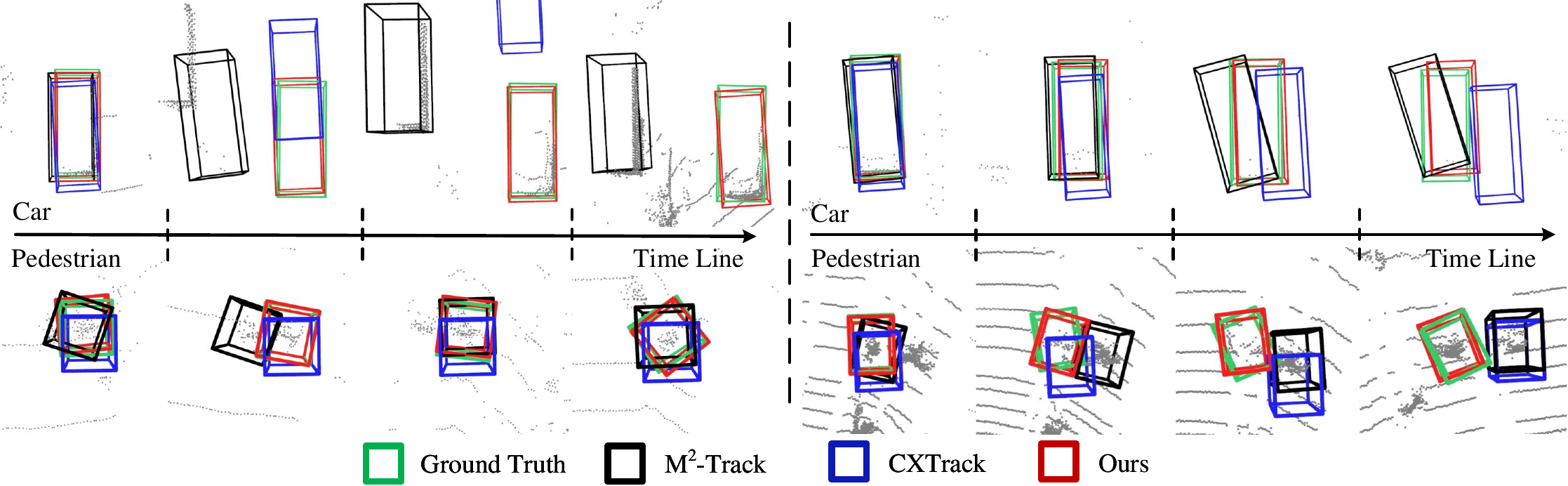}
  \vspace{-0.13in}
	\caption{\textbf{Visualization tracking results on KITTI dataset.} We compare our method with M${^2}$-Track, CXTrack and show the cases of Car and Pedestrian.}
	\label{fig:Quantitative results}
 \vspace{-0.15in}
\end{figure*} 

\begin{table}[t!]
\renewcommand\tabcolsep{3.5pt}
\begin{center}
\vspace{0.1in}
\caption{Performance comparison on the NuScenes dataset. Ped is an abbreviation for pedestrian. {\dag} represents the results we train and test with the official code.}~\label{tab:nuscenes result}
\vspace{-0.13in}
\begin{tabular}{l|c|cccccc}
\toprule[.05cm]
\multirow{2}*{}
& Category & Car &Ped. & Truck & Trailer & Bus & Mean \\
& Frames &\textit{64,159} &\textit{33,227} &\textit{13,587} &\textit{3,352} &\textit{2,953} &\textit{117,278} \\
\hline
\hline
\rule{0pt}{8pt}
\multirow{9}*{\rotatebox{90}{\textit{Success}}}
& SC3D~\cite{SC3D} & 22.31 & 11.29 & 30.67 & 35.28 & 29.35 & 20.70 \\
& P2B~\cite{P2B} & 38.81 & 28.39 & 42.95 & 48.96 & 32.95 & 36.48 \\
& BAT~\cite{BAT} & 40.73 & 28.83 & 45.34 & 52.59 & 35.44 & 38.10 \\
& SMAT~\cite{SMAT} & 43.51 & 32.27 & 44.78 & 37.45 & 39.42 & 40.20 \\
& C2FT~\cite{C2FT} & 40.80 & - & 48.40 & 58.50 & 40.50 & - \\
& MLSET~\cite{MLSET} & 53.20 & \underline{33.20} & 54.30 & 53.10 & - & - \\
& CXTrack{\dag}~\cite{CXTrack} &48.92 & 31.67 &51.40 &\textbf{60.64} &40.11 &44.43 \\
& M$^2$-Track~\cite{beyond} & \underline{55.85} & 32.10 & \underline{57.36} & 57.61 & \underline{51.39} & \underline{49.23} \\
& \textbf{Ours} & \textbf{57.05} & \textbf{37.08} & \textbf{59.37} & \underline{59.73} & \textbf{55.46} & \textbf{51.70} \\
\hline
\hline
\rule{0pt}{8pt}
\multirow{9}*{\rotatebox{90}{\textit{Precision}}}
& SC3D~\cite{SC3D} & 21.93 & 12.65 & 27.73 & 28.12 & 24.08 & 20.20 \\
& P2B~\cite{P2B} & 42.18 & 52.24 & 41.59 & 40.05 & 27.41 & 45.08 \\
& BAT~\cite{BAT} & 43.29 & 53.32 & 42.58 & 44.89 & 28.01 & 45.71 \\
& SMAT~\cite{SMAT} & 49.04 & 60.28 & 44.69 & 34.10 & 34.32 & 50.92 \\
& C2FT~\cite{C2FT} & 43.80 & - & 46.60 & 51.80 & 36.60 & - \\
& MLSET~\cite{MLSET} & 58.30 & 58.60 & 52.50 & 40.90 & - & - \\
& CXTrack{\dag}~\cite{CXTrack} &55.61 & 56.64 &50.93 &54.44 &35.83 &54.83 \\
& M$^2$-Track~\cite{beyond} & \underline{65.09} & \underline{60.92} & \underline{59.54} & \underline{58.26} & \underline{51.44} & \underline{62.73} \\
& \textbf{Ours} & \textbf{65.94} & \textbf{72.80} & \textbf{63.69} & \textbf{60.44} & \textbf{54.31} & \textbf{67.17} \\
\toprule[.05cm]
\end{tabular}
\end{center}
\vspace{-0.3in}
\end{table}

\subsection{Target Regression and Loss Function} 
To obtain target location $(x_{t}, y_{t})$, we suppose that target is located in the center of template frame following the~\cite{P2B,BAT,PTT}. Therefore, we can use center coordinate to index target motion $(\hat{\Delta x}_{t-1}^{t}, \hat{\Delta y}_{t-1}^{t})$ from motion map $M^{N}$ and obtain the current target location through $(x_{t},y_{t}) = (x_{t-1}+\hat{\Delta x}_{t-1}^{t}, y_{t-1}+\hat{\Delta y}_{t-1}^{t})$. 
Then, we utilize search feature $\hat F^{s}$ to predict a z-axis map $Z_{t} \in \mathbb{R}^{H\times L\times 1}$ and an orientation map $\Theta_{t} \in \mathbb{R}^{H\times L\times 1}$. Finally, we exploit target location $(x_{t},y_{t})$ to index the vertical location $z_{t}$ and orientation $\theta_{t}$ from the predicted map.

Our method is trained with box keypoint loss $L_{kp}$, motion refinement loss $L_{mt}$ and regression loss $L_{reg}$ as follows:
\begin{equation}
L_{kp} = \frac{1}{N_{k}} \sum_{n=1}^{N_{k}} (\Delta C_{n} - \hat{\Delta C_{n}})^{2}
\end{equation}
\begin{equation}
L_{mt} = \frac{1}{N_{pos}} \sum_{n=1}^{N}  \mathbb{I} \left | M_{n} - \hat{M_{n}} \right |, \ 
L_{reg} = \frac{1}{N_{pos}} \sum_{n=1}^{N} \mathbb{I} \left | \gamma_{n} - \hat{\gamma_{n}} \right |
\end{equation}
where $\mathbb{I}$ is the indicator function denoting occupation situation, $\gamma_{n}$ is the object orientation and z-axis location. $N_{k}$, $N_{pos}$ and $N$ mean the number of box keypoints, occupied grids and all grids respectively. The whole training loss is the sum of the above losses with special weights.

\section{EXPERIMENTS}
\subsection{Experimental Settings} 
\noindent \textbf{Datasets.} 
We evaluate our proposed approach on the widely used datasets: KITTI~\cite{KITTI} and NuScenes~\cite{NuScenes}. For KITTI, following the previous work~\cite{P2B,BAT,PTT}, we split the scenes 0-16 for training, scenes 17-18 for validation, and scenes 19-20 for testing. For NuScenes with larger data volumes and more target categories, we follow implementation of~\cite{beyond} to divide 850 scenes into training and validation sets.

\begin{table}[t!]
\renewcommand\tabcolsep{7pt}
\begin{center}
\caption{Ablation studies on model components. BMP means Box Motion Predictor. RIM represents Reciprocating Interaction Module. IRM is Iterative Refinement Module.}~\label{tab:ablation study}
\begin{tabular}{c|c|c|c|c|c}
\toprule[.05cm]
BMP & RIM & IRM & Car & Pedestrian & Mean \\ \hline \hline
& \Checkmark   & \Checkmark   & 70.5 / 82.1 & 62.8 / 88.9 & 66.8 / 85.4 \\
\Checkmark   &     & \Checkmark   & 64.9 / 77.6 & 63.8 / 89.5 &  64.4 / 83.4 \\
\Checkmark   & \Checkmark   & \Checkmark   & \textbf{73.1} / \textbf{84.5}  & \textbf{70.4} / \textbf{95.1} & \textbf{71.8} / \textbf{89.7}             \\ 
\toprule[.05cm]
\end{tabular}
\end{center}
\vspace{-0.2in}
\end{table}

\begin{table}[t]
\renewcommand\tabcolsep{2.0pt}
\caption{Ablation studies on Box Motion Predictor. Setting used in our final model are underlined.}
\vspace{-0.1in}
~\label{tab:BMP Ablation}
\begin{tabular}{c|c|c|c|c}
\toprule[.06cm]
Experiment                        & Method           & Car         & Pedestrian  & Mean        \\ \hline \hline
\rule{0pt}{8pt}
\multirow{3}{*}{Motion Predictor} & LSTM~\cite{lstm}             & 71.4 / 83.5 & 65.7 / 92.5 & 68.6 / 87.9 \\
                                  & GRU~\cite{gru}              & 72.3 / 84.0 & 68.7 / 92.7 & 70.5 / 88.2 \\
                                  & \underline{BMP}  & \textbf{73.1} / \textbf{84.5} & \textbf{70.4} / \textbf{95.1} & \textbf{71.8} / \textbf{89.7} \\
                                  \hline
                                  \hline
                                  \rule{0pt}{8pt}
\multirow{3}{*}{Box Keypoints}    & Only Corners     & 71.0 / 82.5 & 69.5 / 93.2 & 70.3 / 87.7 \\
                                  & Only Center      & 71.8 / 83.4 & 70.1 / 93.5 & 71.0 / 88.3 \\
                                  & \underline{Corners + Center} & \textbf{73.1} / \textbf{84.5} & \textbf{70.4} / \textbf{95.1} & \textbf{71.8} / \textbf{89.7} \\ 
\toprule[.06cm]
\end{tabular}
\vspace{-0.17in}
\end{table}

\noindent \textbf{Evaluation Metric.} 
Consistent with present approaches, we employ One Pass Evaluation (OPE) to measure the \textit{Success} and \textit{Precision} for tracking results. \textit{Success} denotes Intersection Over Union (IOU) between ground-truth and predicted box, while \textit{Precision} measures Area Under Curve (AUC) for distance between two box centers ranging from 0 to 2 meters.

\noindent \textbf{Training and Inference Details.}
In data processing, we set the search and template region as [-3.2$m$, 3.2$m$] for the X and Y axes, and [-3$m$, 1$m$] for the Z axis. Following~\cite{lttr, SMAT}, we voxelize point cloud with the voxel size [0.025$m$, 0.025$m$, 0.05$m$]. Note that we employ random offsets to ground truth during training to simulate the historical boxes with errors in testing. For KITTI and nuScenes datasets, we train 40 epochs for our proposed method with the Adam optimizer and an initial learning rate of 0.001. The training procedure spends about 3 hours on a NVIDIA 3090 GPU with batch size 128. During inference, our tracker exploits several historical boxes and two point cloud frames to capture target frame-by-frame, meanwhile running speed achieves 24.7 FPS on single GPU.

\subsection{Comparison with State-of-the-arts} 
\noindent \textbf{Results on KITTI.} 
As displayed in Tab.~\ref{tab:kitti result}, our MTM-Tracker outperforms other methods in all categories. Specifically, our method surpasses previous state-of-the-art method CXTrack \cite{CXTrack} of 3.4\%/3.1\% in the average performance. We speculate that compared with CXTrack which depends on contextual information of point clouds across two frames, our paradigm can overcome occlusion better and relieve interference from the other similar objects due to extra modeling target motion during multiple frames. Moreover, our method surpasses M$^2$-Track~\cite{beyond} significantly ($\uparrow$8.0\%/5.0\%). We think although motion-based M$^2$-Track is better than matching approaches, the inaccurate segmentation and short-term motion modeling may still limit its tracking performance. In contrast, owing to considering long-term motion prior and appearance matching at the same time, our method could locate target accurately. Besides, we exhibit some tracking results in Fig.~\ref{fig:Quantitative results}.

\begin{figure}[t!]
	\centering
	\includegraphics[width=8.3cm,height=2.7cm]{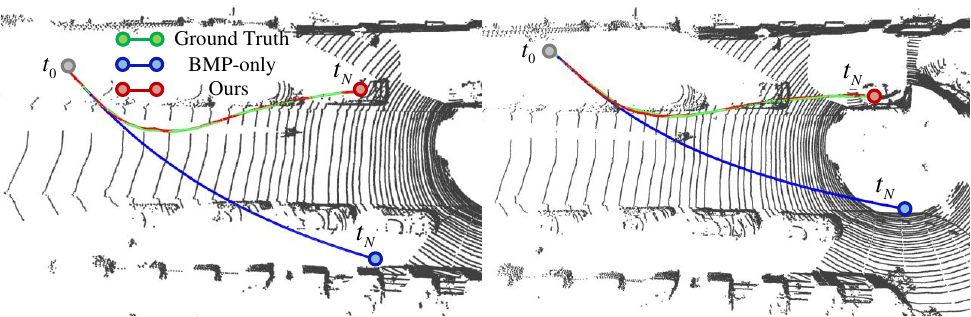}
  \vspace{-0.1in}
	\caption{\textbf{Comparison between BMP-only and MTM-Tracker trajectory.}}
  \vspace{-0.05in}
\label{fig:Only BMP v2}
\end{figure} 

\begin{figure}[t]
	\centering
	\includegraphics[width=8cm,height=3.2cm]{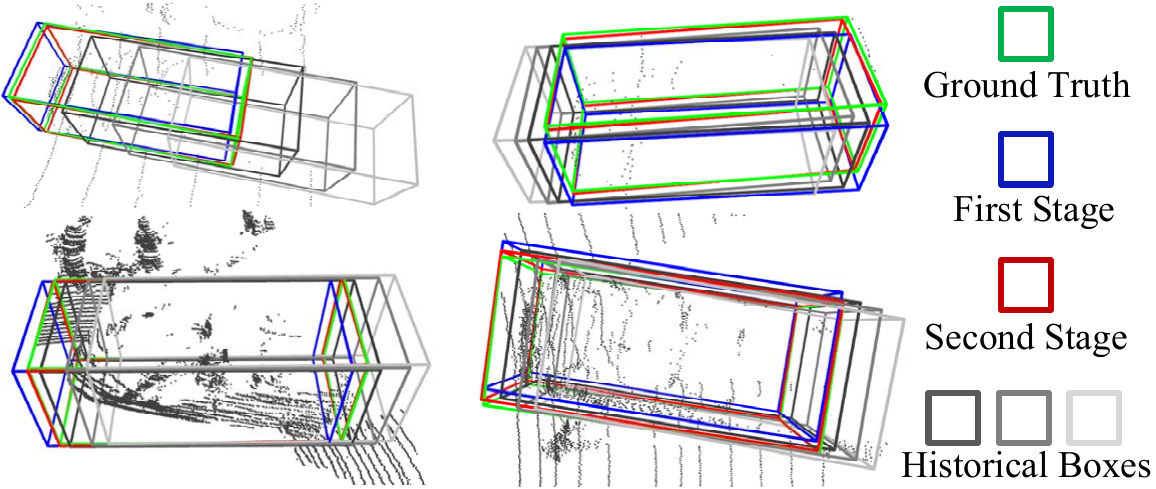}
  \vspace{-0.1in}
	\caption{\textbf{Visualization of prediction refinement.} We present four different objects to show that the refined stage further corrects the coarse predictions.}
	\label{fig:refine}
 \vspace{-0.15in}
\end{figure} 

\noindent \textbf{Results on NuScenes.}
As illustrated in Tab.~\ref{tab:nuscenes result}, our MTM-Tracker obtains the best performance in nuScenes dataset for most categories and demonstrates superior results compared with previous matching paradigm~\cite{SC3D,P2B,BAT,SMAT,C2FT,MLSET}. Specially, the BEV-based method~\cite{SMAT} is hard to regress large objects by directly using CenterHead~\cite{centerpoint} since target center is usually empty. Different from~\cite{SMAT}, our BEV-based MTM-Tracker utilizes motion clues to estimate a coarse center on BEV and further refine it through correlation features. In this way, we not only mitigate the above issue in~\cite{SMAT} but also achieve performance balance among all categories. Notably, our MTM-Tracker surpasses M$^2$-Track~\cite{beyond} in the Pedestrian by 4.98\% \textit{Success} and 11.88\% \textit{Precision}, and obtains certain improvement for the large objects, such as Trailer, Truck and Bus. Meanwhile, we outperform M$^2$-Track by 2.47\%/4.44\% on \textit{Mean} metrics, proving superiority of overall performance in our Motion-to-Matching paradigm.

\begin{table}[t]
\renewcommand\tabcolsep{5pt}
\begin{center}
\vspace{0.05in}
\caption{Comparison of different components in Reciprocating Interaction Module. TD and BU Stream mean top-down and bottom-up feature interaction.}
~\label{tab:frame num}. 
\begin{tabular}{c|c|c|c|c}
\toprule[.05cm]
TD-Stream & BU-Stream & Car & Pedestrian & Mean \\ \hline\hline
\rule{0pt}{8pt}
w/o IAM & \underline{w/ Deform}   & 72.2 / 83.7 & 69.4 / 93.9 & 70.8 / 88.7       \\
w/o Mask & \underline{w/ Deform}  & 68.9 / 79.7 & 68.3 / 93.7 & 68.6 / 86.5       \\
\underline{w/ IAM}  & w/ Concat   & 70.7 / 83.8 & 64.5 / 91.1 & 67.7 / 87.4       \\
\underline{w/ IAM}  & w/ Add & 70.8 / 84.1 & 67.5 / 92.0 & 69.2 / 87.9       \\
\underline{w/ IAM}   & \underline{w/ Deform}  & \textbf{73.1} / \textbf{84.5} & \textbf{70.4} / \textbf{95.1}  & \textbf{71.8} / \textbf{89.7}   \\ \hline
\toprule[.05cm]
\end{tabular}
\vspace{-0.15in}
\end{center}
\end{table}

\begin{table}[]
\renewcommand\tabcolsep{8pt}
\begin{center}
\caption{Ablation studies on multi-scale features in Reciprocating Interaction Module.}
~\label{tab:multi-scale}. 
\begin{tabular}{c|c|c|c}
\toprule[.06cm]
Multi-Scale Layer & Car         & Pedestrian  & Mean        \\ \hline \hline
\rule{0pt}{8pt}
$H_0$ & 70.2 / 80.6 & 67.6 / 91.4 & 68.9 / 85.9            \\
$H_0 \ \& \ H_1$          & 71.3 / 82.4 & 67.8 / 92.9 & 69.6 / 87.5            \\
$H_0 \ \& \ H_1 \ \& \ H_2$ & \textbf{73.1} / \textbf{84.5} & \textbf{70.4} / \textbf{95.1} & \textbf{71.8} / \textbf{89.7} \\ \toprule[.06cm]
\end{tabular}
\vspace{-0.2in}
\end{center}
\end{table}

\subsection{Ablation Studies} 
Based on KITTI dataset, we conduct comprehensive experiments in Car and Pedestrian categories to validate the effectiveness of our proposed method.

\noindent \textbf{Model Components.} 
The $1^{st}$ row in Tab.~\ref{tab:ablation study} shows that when BMP is not used to provide motion prior to the second stage, it will cause some adverse effects. Especially for Pedestrians that are easily confused, this effect is more significant. Then, as demonstrated in the $2^{nd}$ row, the \textit{Mean} performance will drop 7.4\%/6.3\% without feature interaction, which proves the effectiveness of our RIM. Moreover, in Fig.~\ref{fig:Only BMP v2}, we compare the tracking trajectories generated by BMP-only and our two-stage tracker. The result shows that due to the cumulative errors, BMP loses target when it turns, while the two-stage tracker maintains robust tracking since the second stage can exploit point clouds to refine the target states. Finally, Fig.~\ref{fig:refine} shows that the coarse prediction from BMP can be corrected through feature matching in the second stage. 

\noindent \textbf{Box Motion Prediction.} In the case of utilizing the same number of historical boxes, we compare our BMP with other sequence-to-sequence models including the LSTM~\cite{lstm} and GRU~\cite{gru}. As shown in Tab.~\ref{tab:BMP Ablation}, the LSTM and GRU are lower than BMP by 3.2\%/1.8\% and 1.3\%/1.5\% on \textit{Mean} performance. We speculate that due to the appearance of some inaccurate predicted boxes in historical sequence, the unreliable motion information would be encoded into hidden features in the~\cite{lstm,gru} and hinder final motion prediction. However, owing to the attention mechanism, our BMP could suppress negative impacts from low-quality boxes via learnable weights and model global motion pattern better.

\noindent \textbf{Box Keypoints Selection.} In Tab.~\ref{tab:BMP Ablation}, we explore the effect of choosing different box keypoints on tracking performance. Specifically, in the 1$^{st}$ row, lacking explicit center constraint will lead to performance degradation ($\downarrow$1.5\%/2.0\% in average metrics). Besides, in the 2$^{nd}$ row, using continuous center still cannot realize optimal prediction. However, when adding the overall motion provided by box corners, the performance can improve $0.8\%$/$1.4\%$. Thus, we believe that the box center and corners are both indispensable in box motion prediction. 

\begin{table}[t]
\renewcommand\tabcolsep{9pt}
\begin{center}
\vspace{0.05in}
\caption{Effects of different regression strategies on performance.}~\label{tab:regression strategy}
\begin{tabular}{c|c|c|c}
\toprule[.06cm]
Regression Strategy & Car & Pedestrian & Mean \\ \hline\hline
\rule{0pt}{8pt}
$(\Delta x,\Delta y,\Delta z,\Delta \theta)$  & 61.6 / 71.7 &  59.3 / 88.3  & 60.5 / 79.8            \\
$(\Delta x,\Delta y,\Delta z,\theta)$         & 62.2 / 72.5 &  60.7 / 89.0  & 61.5 / 80.5            \\
$(\Delta x,\Delta y,z,\Delta \theta)$         & 66.3 / 78.8 &  67.9 / 93.5  & 67.1 / 86.0            \\
$(\Delta x,\Delta y,z,\theta)$                &\textbf{73.1} / \textbf{84.5}& \textbf{70.4} / \textbf{95.1} & \textbf{71.8} / \textbf{89.7}             \\
\toprule[.06cm]
\end{tabular}
\end{center}
\vspace{-0.2in}
\end{table}

\begin{figure}[t]
\centering
\includegraphics[width=\linewidth]{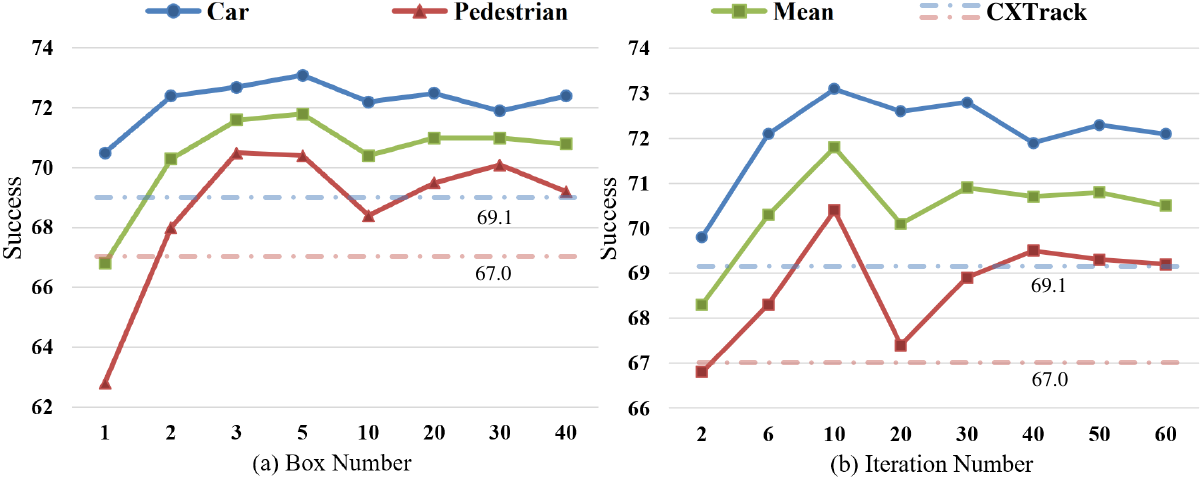}
\vspace{-0.2in}
\caption{\textbf{(a) The performance of varying numbers of historical boxes. (b) Effect of the number of iterations in the Iterative Refinement Module.}}
\label{fig:Iteration Number}
\vspace{-0.18in}
\end{figure} 

\noindent \textbf{Historical Boxes Input.}
As shown in Fig.~\ref{fig:Iteration Number}(a), we study the impact on performance when different numbers of historical boxes are utilized to estimate box motion. When the number increases from one to two, MTM-Tracker achieves significant improvement, especially $\uparrow$5.2\% \textit{Success} in Pedestrian class. Then, the performance will improve steadily until the boxes increase to five. 
After that, feeding too many historical boxes leads to performance fluctuations. We think that a few boxes already provide enough motion clues, such as velocity, while further increasing boxes will bring excessive complexity and redundant information to cause the negative effects in certain.

\noindent \textbf{Reciprocating Interaction.} 
In Tab.~\ref{tab:frame num}, we validate effectiveness of our Reciprocating Interaction Module. As illustrated in the 1$^{st}$ row, the top-down stream without IAM leads to a decrease of 1.0\%/1.0\% in \textit{Mean}, verifying that proposed interactive attention plays an essential role in learning spatial relation. Then, the 2$^{nd}$ row confirms the necessity of using template mask to distinguish target from background. Additionally, from the 3$^{rd}$, 4$^{th}$ and 5$^{th}$ row, we can observe that our deformable operation could achieve the best result due to adaptively aligning multi-scale features and integrating information in deformable manner.
Moreover, Tab.~\ref{tab:multi-scale} shows that benefiting from perceiving target from multi-scale features, our method could catch target at various speeds and achieve better performance.

\noindent \textbf{Iterative Refinement.}
In Fig.~\ref{fig:Iteration Number}(b), we demonstrate the relationship between the number of iterations and performance. We can see that as iteration increases, the performance will first rise obviously and tend to stabilize but be slightly lower than optimal value. We speculate that excessive iteration will cause motion increment in each iteration to be too small and affect the final result. Nonetheless, as shown in Fig.~\ref{fig:Iteration Number}(b), our method still outperforms the previous state-of-the-art method CXTrack~\cite{CXTrack} with remarkable margins in most cases.

\noindent \textbf{Regression Strategy.} 
For the orientation and z-axis location regression, we compare performance when predicting relative and absolute values, and the results are displayed in Tab.~\ref{tab:regression strategy}. First, if using $\Delta z$ and $\Delta \theta$ as regression results which are estimated by BMP and refined by the second stage, the \textit{Mean} will decline 11.3\% and 9.9\%. We think this is because the changes in the z-axis location and orientation of target are usually too small between two frames. Thus, it is difficult to refine relative values accurately and will cause accumulative errors, resulting in deviating from the ground truth. Then, in the 2$^{nd}$ and 3$^{rd}$ row, we replace $\Delta z$ and $\Delta \theta$ with absolute predictions respectively and observe that the z-axis regression strategy has more significant impact on performance. Finally, in the 4$^{th}$ row, the performance can be further improved when the absolute values $z$ and $\theta$ are applied.

\section{Conclusion} 
In this work, we investigate the limitations of existing 3D SOT methods and propose a novel mixed tracking paradigm, which integrates motion modeling with appearance matching via a unified network. Additionally, we propose a two-stage pipeline MTM-Tracker containing motion prediction, feature interaction and motion refinement. The comprehensive experiments prove the effectiveness of our proposed modules and demonstrate that our method has competitive performance in various aspects.

\bibliographystyle{ieeetr}
\bibliography{my_egbib}

\begin{thebibliography}{10}

\bibitem{P2B}
H.~Qi, C.~Feng, Z.~Cao, F.~Zhao, and Y.~Xiao, ``P2b: Point-to-box network for 3d object tracking in point clouds,'' in {\em IEEE/CVF Conference on Computer Vision and Pattern Recognition}, 2020.

\bibitem{BAT}
C.~Zheng, X.~Yan, J.~Gao, W.~Zhao, W.~Zhang, Z.~Li, and S.~Cui, ``Box-aware feature enhancement for single object tracking on point clouds,'' 2021.

\bibitem{PTT}
J.~Shan, S.~Zhou, Z.~Fang, and Y.~Cui, ``Ptt: Point-track-transformer module for 3d single object tracking in point clouds,'' 2021.

\bibitem{lttr}
Y.~Cui, Z.~Fang, J.~Shan, Z.~Gu, and S.~Zhou, ``3d object tracking with transformer,'' 2021.

\bibitem{v2b}
L.~Hui, L.~Wang, M.~Cheng, J.~Xie, and J.~Yang, ``3d siamese voxel-to-bev tracker for sparse point clouds,'' 2021.

\bibitem{C2FT}
B.~Fan, K.~Wang, H.~Zhang, and J.~Tian, ``Accurate 3d single object tracker with local-to-global feature refinement,'' {\em IEEE Robotics and Automation Letters}, vol.~7, no.~4, pp.~12211--12218, 2022.

\bibitem{MLSET}
Q.~Wu, C.~Sun, and J.~Wang, ``Multi-level structure-enhanced network for 3d single object tracking in sparse point clouds,'' {\em IEEE Robotics and Automation Letters}, vol.~8, no.~1, pp.~9--16, 2023.

\bibitem{SiamFC}
L.~Bertinetto, J.~Valmadre, J.~F. Henriques, A.~Vedaldi, and P.~H.~S. Torr, ``Fully-convolutional siamese networks for object tracking,'' in {\em European Conference on Computer Vision Workshops} (G.~Hua and H.~J{\'e}gou, eds.), pp.~850--865, 2016.

\bibitem{SiamRPN}
B.~Li, J.~Yan, W.~Wu, Z.~Zhu, and X.~Hu, ``High performance visual tracking with siamese region proposal network,'' in {\em IEEE/CVF Conference on Computer Vision and Pattern Recognition}, pp.~8971--8980, 2018.

\bibitem{SiamMask}
Q.~Wang, L.~Zhang, L.~Bertinetto, W.~Hu, and P.~H. Torr, ``Fast online object tracking and segmentation: A unifying approach,'' in {\em Proceedings of the IEEE/CVF conference on Computer Vision and Pattern Recognition}, pp.~1328--1338, 2019.

\bibitem{beyond}
C.~Zheng, X.~Yan, H.~Zhang, B.~Wang, S.~Cheng, S.~Cui, and Z.~Li, ``Beyond 3d siamese tracking: A motion-centric paradigm for 3d single object tracking in point clouds,'' in {\em Proceedings of the IEEE/CVF Conference on Computer Vision and Pattern Recognition}, pp.~8111--8120, 2022.

\bibitem{TAT}
K.~Lan, H.~Jiang, and J.~Xie, ``Temporal-aware siamese tracker: Integrate temporal context for 3d object tracking,'' in {\em Proceedings of the Asian Conference on Computer Vision}, pp.~399--414, 2022.

\bibitem{lstm}
S.~Hochreiter and J.~Schmidhuber, ``Long short-term memory,'' {\em Neural computation}, vol.~9, no.~8, pp.~1735--1780, 1997.

\bibitem{gru}
K.~Cho, B.~Van~Merri{\"e}nboer, C.~Gulcehre, D.~Bahdanau, F.~Bougares, H.~Schwenk, and Y.~Bengio, ``Learning phrase representations using rnn encoder-decoder for statistical machine translation,'' {\em arXiv preprint arXiv:1406.1078}, 2014.

\bibitem{siamcar}
D.~Guo, J.~Wang, Y.~Cui, Z.~Wang, and S.~Chen, ``Siamcar: Siamese fully convolutional classification and regression for visual tracking,'' in {\em Proceedings of the IEEE/CVF conference on computer vision and pattern recognition}, pp.~6269--6277, 2020.

\bibitem{siameseban}
Z.~Chen, B.~Zhong, G.~Li, S.~Zhang, and R.~Ji, ``Siamese box adaptive network for visual tracking,'' in {\em Proceedings of the IEEE/CVF conference on computer vision and pattern recognition}, pp.~6668--6677, 2020.

\bibitem{Transformer}
A.~Vaswani, N.~Shazeer, N.~Parmar, J.~Uszkoreit, L.~Jones, A.~N. Gomez, L.~u. Kaiser, and I.~Polosukhin, ``Attention is all you need,'' in {\em Advances in Neural Information Processing Systems}, vol.~30, Curran Associates, Inc., 2017.

\bibitem{transt}
X.~Chen, B.~Yan, J.~Zhu, D.~Wang, X.~Yang, and H.~Lu, ``Transformer tracking,'' in {\em Proceedings of the IEEE/CVF Conference on Computer Vision and Pattern Recognition}, pp.~8126--8135, 2021.

\bibitem{transformermeetstracker}
N.~Wang, W.~Zhou, J.~Wang, and H.~Li, ``Transformer meets tracker: Exploiting temporal context for robust visual tracking,'' in {\em Proceedings of the IEEE/CVF Conference on Computer Vision and Pattern Recognition}, pp.~1571--1580, 2021.

\bibitem{SC3D}
S.~Giancola, J.~Zarzar, and B.~Ghanem, ``Leveraging shape completion for 3d siamese tracking,'' in {\em Proceedings of the IEEE/CVF Conference on Computer Vision and Pattern Recognition}, 2019.

\bibitem{pttr}
C.~Zhou, Z.~Luo, Y.~Luo, T.~Liu, L.~Pan, Z.~Cai, H.~Zhao, and S.~Lu, ``Pttr: Relational 3d point cloud object tracking with transformer,'' in {\em CVPR}, pp.~8531--8540, 2022.

\bibitem{votenet}
C.~R. Qi, O.~Litany, K.~He, and L.~J. Guibas, ``Deep hough voting for 3d object detection in point clouds,'' in {\em Proceedings of the IEEE/CVF International Conference on Computer Vision}, 2019.

\bibitem{SMAT}
Y.~Cui, J.~Shan, Z.~Gu, Z.~Li, and Z.~Fang, ``Exploiting more information in sparse point cloud for 3d single object tracking,'' {\em IEEE Robotics and Automation Letters}, vol.~7, no.~4, pp.~11926--11933, 2022.

\bibitem{stnet}
L.~Hui, L.~Wang, L.~Tang, K.~Lan, J.~Xie, and J.~Yang, ``3d siamese transformer network for single object tracking on point clouds,'' {\em arXiv preprint arXiv:2207.11995}, 2022.

\bibitem{second}
Y.~Yan, Y.~Mao, and B.~Li, ``Second: Sparsely embedded convolutional detection,'' in {\em Sensors}, vol.~18, 2018.

\bibitem{DeformableDETR}
X.~Zhu, W.~Su, L.~Lu, B.~Li, X.~Wang, and J.~Dai, ``Deformable detr: Deformable transformers for end-to-end object detection,'' 2021.

\bibitem{raft}
Z.~Teed and J.~Deng, ``Raft: Recurrent all-pairs field transforms for optical flow,'' in {\em Computer Vision--ECCV 2020: 16th European Conference, Glasgow, UK, August 23--28, 2020, Proceedings, Part II 16}, pp.~402--419, Springer, 2020.

\bibitem{CXTrack}
T.-X. Xu, Y.-C. Guo, Y.-K. Lai, and S.-H. Zhang, ``Cxtrack: Improving 3d point cloud tracking with contextual information,'' in {\em Proceedings of the IEEE/CVF Conference on Computer Vision and Pattern Recognition}, pp.~1084--1093, 2023.

\bibitem{KITTI}
A.~Geiger, P.~Lenz, and R.~Urtasun, ``Are we ready for autonomous driving? the kitti vision benchmark suite,'' in {\em IEEE Conference on Computer Vision and Pattern Recognition}, pp.~3354--3361, 2012.

\bibitem{NuScenes}
H.~Caesar, V.~Bankiti, A.~H. Lang, S.~Vora, V.~E. Liong, Q.~Xu, A.~Krishnan, Y.~Pan, G.~Baldan, and O.~Beijbom, ``nuscenes: A multimodal dataset for autonomous driving,'' in {\em IEEE/CVF Conference on Computer Vision and Pattern Recognition}, 2020.

\bibitem{centerpoint}
T.~Yin, X.~Zhou, and P.~Krahenbuhl, ``Center-based 3d object detection and tracking,'' in {\em Proceedings of the IEEE/CVF conference on computer vision and pattern recognition}, pp.~11784--11793, 2021.

\end{thebibliography}

\end{document}